\documentclass[sigconf]{acmart}

\AtBeginDocument{%
  }

\setcopyright{acmlicensed}
\copyrightyear{2018}
\acmYear{2018}
\acmDOI{XXXXXXX.XXXXXXX}

\acmConference[Conference acronym 'XX]{Make sure to enter the correct
  conference title from your rights confirmation emai}{June 03--05,
  2018}{Woodstock, NY}
\acmISBN{978-1-4503-XXXX-X/18/06}

\usepackage{bbm}
\usepackage{subfigure}
\usepackage{graphicx}

\begin{document}

\title{No One Left Behind: How to Exploit the Incomplete and Skewed Multi-Label Data for Conversion Rate Prediction}




\author{Qinglin Jia}
\email{jiaql@pku.edu.com}
\authornote{Work done while at Huawei.}
\affiliation{%
  \institution{Peking University}
  \country{China}
}

\author{Zhaocheng Du}
\email{zhaochengdu@huawei.com}
\affiliation{%
  \institution{Noah's Ark Lab, Huawei}
  \country{China}
}

\author{Chuhan Wu}
\email{wuchuhan1@huawei.com}
\affiliation{%
  \institution{Noah's Ark Lab, Huawei}
  \country{China}
}

\author{Huifeng Guo}
\authornote{Corresponding author.}
\email{huifeng.guo@huawei.com}
\affiliation{%
  \institution{Noah's Ark Lab, Huawei}
  \country{China}
}

\author{Ruiming Tang}
\email{tangruiming@huawei.com}
\affiliation{%
  \institution{Noah's Ark Lab, Huawei}
  \country{China}
}

\author{Shuting Shi}
\author{Muyu Zhang}
\email{{shishuting2,zhangmuyu}@huawei.com}
\affiliation{%
  \institution{Huawei Technology Co., Ltd.}
  \country{China}
}
\renewcommand{\shortauthors}{Jia et al.}

\begin{abstract}
  In most real-world online advertising systems, advertisers typically have diverse customer acquisition goals.
  A common solution is to use multi-task learning (MTL) to train a unified model on post-click data to estimate the conversion rate (CVR) for these diverse targets.
  In practice, CVR prediction often encounters missing conversion data as many advertisers submit only a subset of user conversion actions due to privacy or other constraints, making the labels of multi-task data incomplete.
  If the model is trained on all available samples where advertisers submit user conversion actions, it may struggle when deployed to serve a subset of advertisers targeting specific conversion actions, as the training and deployment data distributions are mismatched.
  While considerable MTL efforts have been made, a long-standing challenge is how to effectively train a unified model with the incomplete and skewed multi-label data.

  In this paper, we propose a fine-grained \textbf{K}nowledge transfer framework for \textbf{A}symmetric \textbf{M}ulti-\textbf{L}abel data (KAML). We introduce an attribution-driven masking strategy (ADM) to better utilize data with asymmetric multi-label data in training. 
  However, the more relaxed masking in ADM is a double-edged sword: it provides additional training signals but also introduces noise due to skewed data.
  To address this, we propose a hierarchical knowledge extraction mechanism (HKE) to model the sample discrepancy within the targe task tower.
  Finally, to maximize the utility of unlabeled samples, we incorporate ranking loss strategy to further enhance our model. 
  The effectiveness of KAML has been demonstrated through comprehensive evaluations on offline industry datasets and online A/B tests, which show significant performance improvements over existing MTL baselines.
   

\end{abstract}




\maketitle

\section{INTRODUCTION}
Online advertising plays a critical role in customer acquisition, typically following a sequential event pattern of ``impression $\rightarrow$ click $\rightarrow$ conversion''\cite{OCPC,Apppromotion}.
Advertising platforms must align with user interests to maximize advertisers' return on investment (ROI).
Therefore, post-click conversion rate predictions (CVR)\cite{advertising} are essential tasks for personalized ad ranking systems, as they help optimize advertisers' bid strategies under various pricing models, such as optimized cost-per-click (OCPC) and cost-per-action (CPA) advertising. 
Advertising platforms rely critically on accurate CVR predictions to adjust bidding strategies and optimize budget allocation, thereby driving the effectiveness of online advertising.

Advertisers have diverse customer acquisition needs, which correspond to different post-click conversion behaviors. 
In typical app promotion campaigns, for example, five main customer acquisition goals are associated with distinct conversion actions: \textit{activation}, re-\textit{engagement}, \textit{registration}, \textit{payment}, and \textit{retention}. 
In OCPC advertising, advertisers select a conversion action for bidding based on their customer acquisition goals and submit the target conversion action of users who clicked the ad to the platform, in accordance with the platform's rules. 
This enables the platform to optimize various algorithms, including CVR models, and improve ad delivery, ensuring that ads are presented to users with the highest probability of executing the target conversion action.

Conventional methods build a separate model for each conversion actions \cite{rosales2012post, autoint, final, deepfm},  which can be hindered by conversion data sparsity. Additionally, maintaining multiple separate models requires significant manpower and computing resources, making this approach neither scalable nor efficient. 
To address these challenges, multi-task learning (MTL) has emerged as a popular solution in both industry and academia\cite{likemtl, surveyMTL, chen2020just, yang2023adatask}.
These widely-used multi-task methods involves building a unified model trained on multiple task labels, allowing it to make predictions for all tasks.

\begin{figure}[t]
  \centering
  \includegraphics[width=0.9\linewidth]{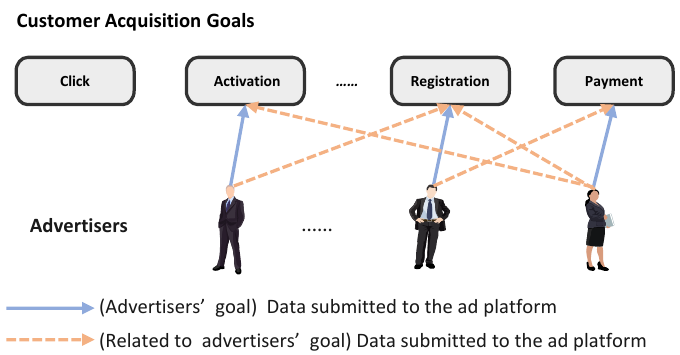}
    \vspace{-0.05in}
  \caption{Advertisers have diverse customer acquisition goals. Each advertiser submits a specific user conversion action that correspond to the customer acquisition goal, and may submits some other user conversion actions.}
  \label{exp}
  \vspace{-0.1in}
\end{figure}

However, in practice, conversion attribution in advertising delivery relies heavily on data provided by advertisers.
After a user clicks on an ad, they are redirected to the advertiser's landing page, and the ad platform is unable to track subsequent user actions.
Due to concerns over user privacy and the need to protect business secrets, most advertisers only submit a portion of user conversion actions, which can lead to incomplete attribution data and referred to as \textbf{asymmetric multi-label data} (Fig.~\ref{exp}). 
In this context, only some of the multiple conversion labels for each post-click sample are accurate. 
For each conversion attribution label in a post-click sample, a label of 1 signifies that user feedback has been observed and confirmed, indicating a successful conversion event.
A label of 0 indicates either that no user action corresponding to the conversion was observed or that the advertiser has not provided sufficient attribution data for that particular click.
This presents a challenge for training multi-task models, as a label of 0 does not necessarily represent a negative sample, which would be misleading and could compromise model accuracy when training.
A straightforward solution is to focus on the specific conversion target relevant to each advertiser’s customer acquisition goal and filter out the irrelevant samples regardless of their labels when training the model on each post-click sample.
In this approach, we consider only the task associated with the targeted conversion for each click, ignoring irrelevant conversion tasks. 
This ensures that the model is trained on the most relevant data, improving the accuracy and relevance of its predictions.

Most multi-task learning (MTL) methods follow this straightforward process, but they fail to fully leverage all available data. 
Motivated by the limitations of the existing MTL approaches, we propose a fine-grained \textbf{K}nowledge transfer framework for \textbf{A}symmetric \textbf{M}ulti-\textbf{L}abels data (KAML).
This framework is designed to train a unified model that maximizes the use of all submitted data, ensuring that no information is left unused.
To achieve this, we introduce an attribution-driven masking strategy (ADM) that generates masks for each post-click sample based on the historical data submitted by advertisers. 
Using these masks, the model identifies both negative and unlabeled samples, and we train the MTL model for each task with binary cross-entropy loss(BCE), considering only positive and negative samples while excluding unlabeled samples. 
However, this approach may still encounter noise due to data skew. Specifically, for each conversion target, the model is trained on all available data from advertisers targeting that conversion, as well as from other advertisers. 
When deployed, the model serves advertisers who are specifically targeting the task. 
Due to the mismatch between training and deployment data, which violates the independent and identically distributed (i.i.d.) assumption — a key principle that assumes training and test data come from the same distribution and are independently sampled — this can introduce bias into the model's predictions.
To alleviate this mismatch, we propose a hierarchical knowledge extraction mechanism (HKE) to model the distribution discrepancy within the target task tower. 
In this approach, samples from advertisers targeting that task and samples from other advertisers are modeled using different parameters, enabling the model to better adapt to the deployment scenario.
To maximize the value of the unlabeled samples, we combine BCE loss with Ranking-based Label Utilization(RLU) strategy to mine information from unlabeled samples.
This approach can not only enhances the model’s ranking ability but also improves its classification ability.

We summarize the major contributions of this paper as follows:
\begin{itemize}
    \item We identify a key challenge in online advertising, stemming from incomplete and skewed multi-label data.
    Specifically, the asymmetry in multi-label data, caused by the partial submission of user conversion actions by advertisers, complicates the training of multi-task models.
    \item We propose a novel Multi-Task Learning model, KAML, designed to fully leverage all available data.
     The attribution-driven masking strategy (ADM) identifies unlabeled samples, while the hierarchical knowledge extraction mechanism (HKE) addresses sample discrepancies across different advertisers.
     Additionally, the Ranking-based Label Utilization (RLU) strategy maximizes the value of unlabeled samples, further mitigating the sparse data problem and enhancing both the ranking and classification performance of the model.
    \item Extensive experiments on offline industry and public datasets and online A/B tests demonstrate the superiority of our method. Deployed on a mainstream 
    online advertising platform, KAML achieved a 12.11\% increase in Revenue Per Mille (RPM) and a 0.92\% improvement in Conversion Rate (CVR), proving its effectiveness in real-world applications.
\end{itemize}

\section{PRELIMINARY}
\begin{figure*}[htbp]
  \centering
  \includegraphics[width=0.8\linewidth]{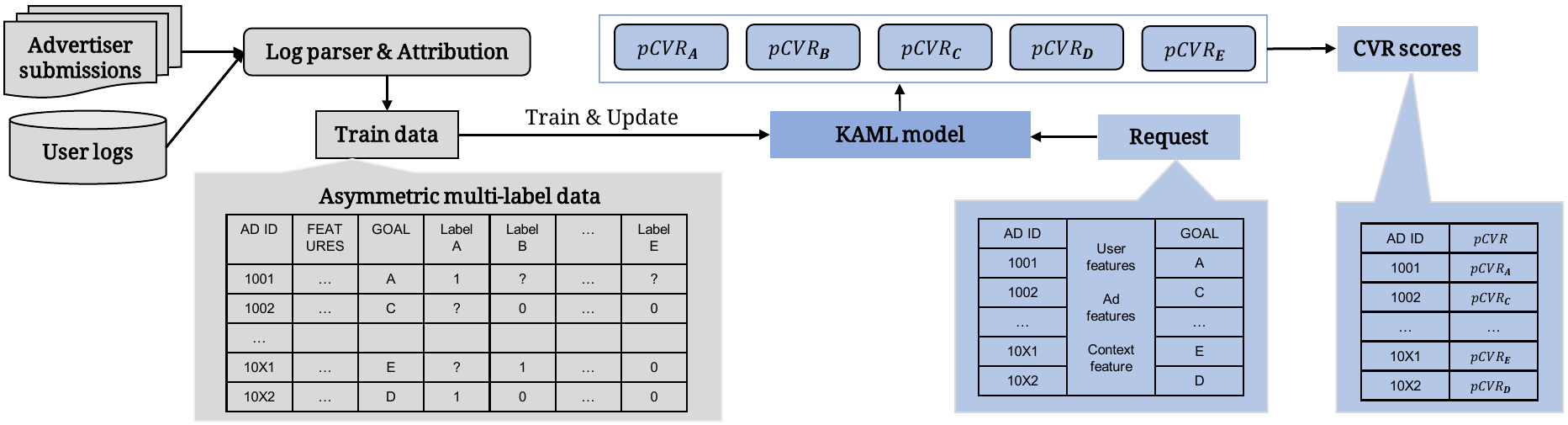}
    \vspace{-0.05in}
  \caption{An MTL-based CVR ranking system for
app promotion in online advertising}
  \label{fig:fig2}
  \vspace{-0.1in}
\end{figure*}


\subsection{Probelm Formulation}


Considering a dataset $\{(x_i, o_i, <y_i^1, y_i^2, ... y_i^N>)\}_{i=1}^{|\mathcal{D}|}$, where $x_i$ represents the feature set, typically comprising user features, ad features, and context features.
$o_i \in \{1, 2, .. N\}$ denotes the target-specific conversion action associated with the post-click sample, such as \textit{Activation}, \textit{Registration}, \textit{Payment}, or other customer acquisition goals pursued by advertisers. This target conversion action determines the specific objective for the ad campaign and guides the optimization of bidding strategies. 
The vector $<y_i^1, y_i^2, ... y_i^N> \in \{0, 1\}$ represents the conversion labels of $N$ conversion events, where each $y_i^j$ corresponds to a binary label indicating whether a specific conversion action has occurred for the given sample. 
In the context  of asymmetric multi-label CVR prediction, advertisers may submit multiple conversion actions, not necessarily limited to the one they are specifically targeting.
For example, an advertiser focused on \textit{Registration} events may also provide \textit{Activation} signals to the platform, contributing to additional conversion labels.
The dataset is thus multi-label, meaning each sample can have multiple associated conversion labels, some of which may be observed and others may be missing or unobserved. A label of $1$ signifies a successful conversion, while a label of $0$ indicates either no conversion occurred or the advertiser did not provide sufficient data for that event. This results in asymmetric multi-label data, where some conversion events are absent or incomplete, posing challenges for training models to predict across diverse conversion targets effectively.

The goal of asymmetric multi-labels CVR prediction is to learn a prediction model $f$ with parameter $\theta$, which minimizes the generalization error over the joint distribution of features, labels, and conversion targets $(x, y,o)\sim(X, Y, O)$:
\begin{equation}
    \mathbb{E}_{(x,y,o)\sim(X,Y,O)} [l(x,y^o;f_{\theta}(x))] ~, 
\end{equation}
where $l$ denotes the loss function and $y^o$ represents the label of the targeted conversion event defined by the advertiser. 
When evaluation, only the labels corresponding to the target conversion actions are considered, which is closely aligned with advertisers' expectations.





\subsection{An MTL-Based CVR Ranking System for App Promotion in Online Advertising}
In this subsection, we will briefly introduce the MTL-Based CVR ranking system deployed in 
our advertising system, which provides superior services to hundreds of millions of users.
As illustrated in Figure \ref{fig:fig2}, the system is designed to handle multiple objectives by modeling different user conversion actions in a MTL framework.
In the offline training process,  we incorporate user click data extracted from user logs and advertiser submissions to train the MTL ranking model for multiple conversion rate (CVR) predictions.
Since different advertisers have distinct customer acquisition goals, the system must tailor its predictions to each advertiser’s specific needs.
Otherwise, it leads to wasted resources, such as computing and traffic resources. 
Therefore, there is a selector in our system to output the prediction score of different conversion actions for different advertisers.
For each online request, the ranking model generates predictions for the targeted conversion actions. These CVR predictions, along with click-through rate (CTR) predictions, are then passed to other modules within the ad ranking system for further processing.
Finally, the ad ranking system makes a decision on ad exposure based on the score. This end-to-end process ensures that ads are both relevant to users and aligned with advertisers’ objectives, ultimately improving the overall effectiveness of the advertising platform.

\section{METHODOLOGY}
In this section, we first provide an overview of an existing model that is commonly adopted for CVR predictions in the industry. 
Subsequently, we present the detailed design of KAML, which is intended to address the aforementioned issue.

\subsection{Base Model}
To better illustrate the motivation and design of our method, we begin by introducing a base model commonly used in CVR prediction, which is widely adopted in the industry. 
It's a common practice to model multiple tasks or different scenarios together with one unified model \cite{PLE,mask}. 
In our base model, the predictions for different conversion events are treated as MTL problems, where each task corresponds to a specific conversion action.

\begin{figure}[t]
  \centering
  \includegraphics[width=0.9\linewidth]{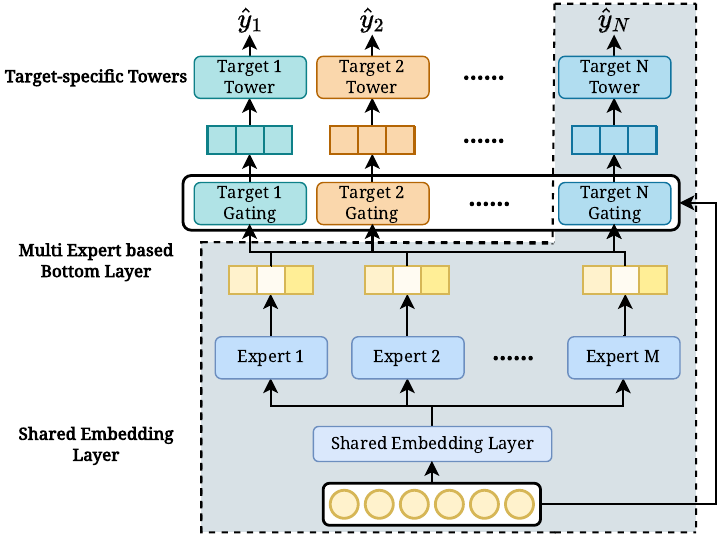}
    \vspace{-0.05in}
  \caption{The architecture of the base model.}
  \label{fig:backbone_model}
  \vspace{-0.1in}
\end{figure}
As shown in Figure \ref{fig:backbone_model}, the base model consists of three parts: Shared Embedding Layer, Multi Experts based Bottom Layer and Target-specific Towers. 
To fully leverage the available training samples and facilitate information sharing across tasks, the base model employs a shared embedding layer that maps the input features into a higher-dimensional embedding space. These embeddings are then concatenated to form the feature representation $x$.

After that, without loss of generality, an Multi-gate Mixture-of-Experts (MMoE-style)\cite{mmoe} bottom layer is designed to capture the similarities and differences among the tasks:
\begin{align}
    g_j(x) = \text{Softmax}(W_j x), \\
    h_j= \sum_{k=1}^{M} g_j^k(x) f_k(x) ~,
\end{align}
where $M$ represents the number of experts, $j$ and $k$ are the indices of the tasks and experts, respectively.
$W_j$ is the weight matrix  of the $j$-th gating network,
$f_k$ is the output of the $k$-th expert,
and $h_j$ is the task-specific representations to obtain prediction results for the $j$-th task.
It is worth mentioning that other common multi-task learning frameworks could also be adopted and replace MMoE here such as PLE\cite{PLE}, HMOE\cite{li2020improving}, TAML\cite{TAML}.

Finally, for each task, the task-specific representations $h_j$ are passed through a Multi-Layer Perceptron (MLP)-based Task-specific Tower to generate predictions for each task:
\begin{equation}
    \hat{y}_j = \text{Sigmoid}(\text{MLP}(h_j)), j \in \{1..N\}~.
\end{equation}

\subsection{Attribution Driven Masking}
As previously discussed, the base model focuses solely on the specific conversion action corresponding to the advertiser's customer acquisition goal.
The loss of the model could be realized in a mask style:

\begin{align}
    \mathcal{L}_{\text{base}} = \sum_{i}^{|\mathcal{D}|} \sum_{j}^{N} \text{Mask}_{ij}^{\text{base}} [y_j &\log(\hat{y}_j) + (1-y_j)\log(1-\hat{y}_j)] ,\\
    \text{Mask}_{ij}^{\text{base}} &= \mathbbm{1}(o_i=j) ~,
\end{align}
where $i$ refers to the index of the sample, and $j$ represents the index of the task.
The mask is non-zero \textbf{if and only if} the task matches the target conversion action. It means that an \textit{Activation} conversion signal submitted by a \textit{Registration}-targeted advertiser will not be utilized for training the \textit{Activation} tower, despite the fact that it did happen and was submitted by the advertiser.


Although such a conservative strategy filters out the noisy signals and enhances the stability of predictions, it results in the loss of a substantial amount of valuable information.
In the context of CVR predictions, where positive samples are scarce, it is a critical challenge to employ these noisy signals in a reasonable manner and the key to improve the prediction ability of the model.

Prior to introducing our methodology, here we supplement some details for online advertising platform. Typically, an advertiser will set several advertising tasks $e_i$ on the platform, where the advertising task is corresponding to a conversion action $o_i$, a promoted product and several creatives. The advertising task provides more macro-level statistical information including conversion rate, CVR prediction bias and other relevant data.
\begin{figure}[bp]
  \centering
  \includegraphics[width=0.9\linewidth]{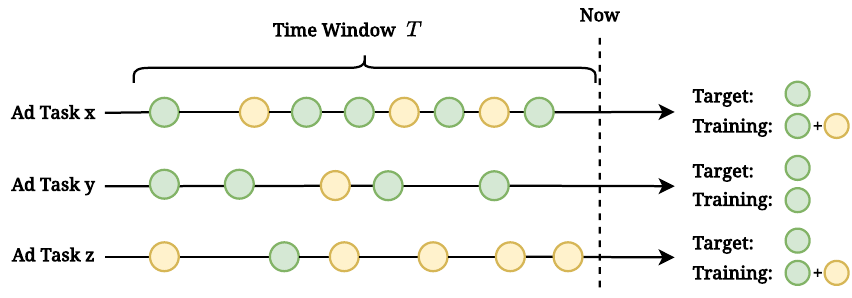}
    \vspace{-0.1in}
  \caption{An exaple of Attribution Driven Masking. The green and yellow circles represent a conversion event of conversion Action A and convertion Action B, respectively.}
  \label{fig:adm_count}
  \vspace{-0.1in}
\end{figure}

In this paper, we propose a novel mask strategy named Attribution Driven Masking(ADM). In contrast to the base model, our objective is to filter the reliable signals from all conversion events provided by advertisers. Specifically, we count the conversion numbers $c_i$ of each advertising task $e_i$ within a time window $T$ and set the mask as the condition if the conversion number $c_i$ is larger than a threshold $\alpha$:
\begin{align}
    \mathcal{L}_{\text{BCE}} = -\frac{1}{|\mathcal{D}|}\sum_{i}^{|\mathcal{D}|} \sum_{j}^{N} \text{Mask}_{ij}^{\text{ADM}} &[y_j \log(\hat{y}_j) + (1-y_j)\log(1-\hat{y}_j)] , \\
    \text{Mask}_{ij}^{\text{ADM}} &= \mathbbm{1}(c_i^j\geq \alpha_j) ~,
\end{align}
where we abuse the notations slightly: The term $c_i^j$ represents the conversion numbers of the conversion target $j$ associated with the corresponding advertising task of the $i$-th sample. $\alpha_j$ is a hyper-parameter controlling the threshold of the conversion target $j$.

To better understand ADM, an example is provided in Figure \ref{fig:adm_count}. For Ad Task $x$, though its target conversion action is Action A, the advertiser submits conversion events of Action B with a quantity exceeding the threshold. Consequently, the samples from Ad Task $x$ will be used for training Action A and B. Conversely, the samples of Task $y$ will only be utilized for training Action A since the submitted conversion events are inadequate and considered unreliable. For Ad Task $z$, though there are hardly any conversion events for Action A, its conversion target is Action A. So samples from Ad Task $z$ will be employed for training Action A and B.

Such a filtering strategy takes into account the historical context of advertising tasks and ensures the reliability of the labels employed in the training process.

\subsection{Hierarchical Knowledge Extraction}
With ADM, a wider range of conversion signals and samples are used for training. However, it should be noted that there is a discrepancy between the distributions of features and labels in the original samples and those extended by ADM. So it is crucial to address this discrepancy and mitigate the effects of negative transfer.

\begin{figure}[t]
    \centering
    \includegraphics[width=0.9\linewidth]{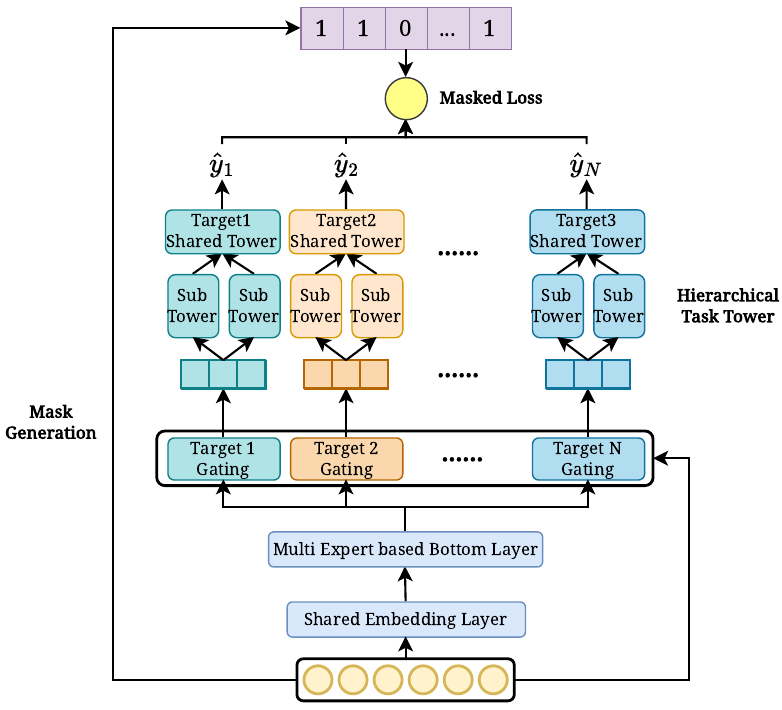}
      \vspace{-0.05in}
    \caption{The architecture of HKE. Two types of samples are routed to two different sub towers.}
    \label{fig:main_model}
    \vspace{-0.1in}
  \end{figure}

As illustrated in Figure \ref{fig:main_model}, we design a hierarchical knowledge extraction mechanism (HKE) to relieve the problem mentioned above. As the name suggests, it extracts the knowledge from the samples of different conversion targets in a hierarchical manner: Firstly, the bottom layer captures the knowledge from different samples with the mask and MMoE mechanism. After that, upon the bottom layer, two sub-towers are employed here to capture the representations derived from the original samples and extended samples separately:
\begin{align}
    h_j^{\text{t}} = MLP_{\text{t}}(h_j), j\in \{1..N\}, \text{t} \in \{\text{original}, \text{extended}\} ~,
\end{align}
where $j$ is the index of the task.
Finally, we merge them and get the predictions of each tasks:
\begin{align}
    \hat{y}_j = \sigma(\text{MLP}(M h_j^{\text{original}} + (1-M) h_j^{\text{extended}})), j \in \{1..N\} ~,
\end{align}
where the indicator $M$ is used to distinguish between the original sample identified by its conversion target and the extended sample by ADM and $\sigma$ is the Sigmoid function.

The HKE method allows for the capture of both shared and specific information between the two types of samples, thereby enhancing the overall framework's performance.

\subsection{Utilizing Unlabeled Samples with Ranking Loss}
The incorporation of ADM and HKE enables the utilization of a more expansive and dependable conversion signal. Nevertheless, further signals could be extracted from the training samples with asymmetric multi-labels. In this part, we present a further strategy for utilizing labels named Ranking-based Label Utilization (RLU).

\begin{figure}[t]
    \centering
    \includegraphics[width=0.85\linewidth]{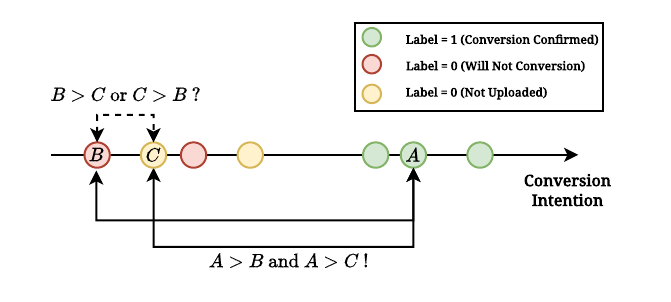}
      \vspace{-0.05in}
    \caption{The comparison of conversion intention among the different samples. Dashed and solid arrows represent uncertain and certain relationships, respectively.}
    \label{fig:ranking_loss}
    \vspace{-0.1in}
  \end{figure}

As shown in Figure \ref{fig:ranking_loss}, the samples in the training set could be divided into three types by their labels: 
\begin{itemize}
    \item \textbf{Type A}. $y=1$. These samples are with confirmed labels uploaded by the advertisers.
    \item \textbf{Type B}. $y=0$ and these samples will not be converted in the end.
    \item \textbf{Type C}. $y=0$ but may be converted actually. The absence of conversion labels on these samples is just because the advertisers did not upload due to the discrepancy between the conversion event and its conversion targets, delayed feedback and other reasons, such as privacy concerns.
\end{itemize}

The samples were positioned on the axis according to the specified conversion intention, as illustrated in Figure \ref{fig:ranking_loss}. The comparisons between the Type B and Type C samples on the conversion intention are invalid, since we could not distinguish the fake negative samples and the real negative samples. However, for Type C samples, conversion intention is evidently lower than that of Type A samples even though their real labels are unknown.

With the relationships of conversion intention, we incorporate a pair-wise ranking loss\cite{burges2005learning} into our framework to supplement more information:
\begin{equation}
    \mathcal{L_{\text{Ranking}}} = - \frac{1}{|\mathcal{D}|}\sum_{i,j, \forall i\neq j}^{|\mathcal{D}|} \sum_{k}^N \beta_k \cdot \mathbbm{1}(y^i_k>y^j_k)\log \frac{1}{1+e^{-(s^i_k-s^j_k)}} ~,
\end{equation}
where $i, j$ are the indices of training samples and $k$ is the index of tasks. $\beta_k$ is the weight of ranking loss for task $k$. $s_k^i, s_k^j$ denote the logits of $i$-th and $j$-th sample for task $k$, respectively. The indicator serves to filter out the uncertain comparisons from the total ranking loss and thus reducing the noises in the labels.

\subsection{Joint Optimization for KAML}
Due to the adaptive mask strategy, the number of valid samples (with non-zero mask) corresponding to the tasks varies among the batches. To make the process of training more stable, like \cite{mask}, we adopt a dynamic average loss on the all valid samples:

\begin{equation}
    \bar{\mathcal{L}}_{\text{BCE}} = - \sum_{j}^{N} \frac{1}{|\sum \text{Mask}_j|}\sum_{i}^{|\mathcal{D}|}  \text{Mask}_{ij}^{\text{ADM}} [y_j \log(\hat{y}_j) + (1-y_j)\log(1-\hat{y}_j)] ~.
\end{equation}

Compared with the base model, KAML improves from the three perspectives: masks, model architectures and losses. The complete loss of KCTL is comprised of two parts: 
\begin{equation}
    \mathcal{L}_{\text{all}} =  \gamma \bar{\mathcal{L}}_{\text{BCE}} + (1-\gamma) \mathcal{L}_{\text{Ranking}} ~,
\end{equation}
where $\gamma \in [0, 1]$ is the hyper-parameter to balance two losses.
\section{EXPERIMENT}

\subsection{Experimental Settings}
\subsubsection{Dataset} 
In the absence of a publicly available dataset that is suitable for the study of the asymmetric multi-label problem, 
we use our online advertising production data to perform the primary offline evaluation.
The dataset includes five conversion actions related to the app promotion business.
Due to confidentiality requirements, we are unable to disclose specific details such as the positive sample ratios. Instead, we refer to these conversion actions as Action A through Action E.
The distribution of data across each conversion action is presented in the "Base" row of Table \ref{tab:dataset_count}.
The training set consists of data collected over a four-week period, with data from the following day used for testing.
The total size of the dataset is approximately $2 \times 10^8$ samples.

To further validate the effectiveness of our model, we also adapt the public KuaiRand-Pure dataset \cite{gao2022kuairand} into a format suitable for asymmetric multi-label data.
KuaiRand is a dataset based on the short-video interactions and comprises multiple types of feedback. We select 4 types feedbacks (\textit{is\_like, is\_follow, is\_comment, is\_forward}
\footnote{For convenience, the feedbacks are also represented as conversion action A to D.}) among all of the 12 feedbacks and regard them as different conversion actions in the industrial dataset.
For each short video, only one type of feedback is randomly selected as the original target conversion action.
The dataset contains approximately  $9 \times 10^5$ samples, with the training set and the test set divided by timestamps in a proportion of 8:2. 
To simplify the training process, negative sampling is performed on samples where all labels are zero, and all dense features are removed.

\begin{table}[t]
    \centering
    \caption{The proportion of training samples used for each task. For the base mask strategy, samples with Conversion Action "Other" will not be used for training.}
    \resizebox{\linewidth}{!}{
    \begin{tabular}{ccccccc}
    \toprule
        Mask Strategy & Action A & Action B &
        Action C &
        Action D &
        Action E &
        Other \\
        \cmidrule{1-7} Base & 69.08\% & 5.26\% & 9.92\% & 11.48\% & 0.56\% & 3.70\%\\
        With ADM & 98.50\% & 7.93\% & 23.34\% & 31.38\% & 53.18\% & -\\
    \bottomrule
    \end{tabular}
    }
    \label{tab:dataset_count}
    \vspace{-0.1in}
\end{table}

\begin{table*}[htbp]
    \centering
    \caption{Results on the industrial dataset. For each row, the best results and the second best results are identified by the bold font and underlining, respectively.}
    \resizebox{0.95\linewidth}{!}{
    \begin{tabular}{ccccccccccccc}
    \toprule
        Conversion Action & \multicolumn{2}{c}{Action A} & \multicolumn{2}{c}{Action B} &
        \multicolumn{2}{c}{Action C} &
        \multicolumn{2}{c}{Action D} &
        \multicolumn{2}{c}{Action E} &
        \multicolumn{2}{c}{All} \\
        \cmidrule{1-13} Metric & AUC & LogLoss & AUC & LogLoss & AUC & LogLoss& AUC & LogLoss& AUC & LogLoss& AUC & LogLoss\\
        \midrule
        SingleTask & 0.8498 & 0.3741 & 0.8921 & 0.4033 & 0.9102 & 0.1738 & 0.8177 & 0.1308 & 0.9038 & 0.0682 & - & - \\ 
        \midrule
        MMoE & 0.8448 & 0.3787 & 0.8921 & 0.4024 & 0.9139 & 0.1706 & 0.8300 & 0.1275 & 0.9061 & 0.0675 & 0.9108 & 0.2469 \\ 
        SharedBottom & 0.8446 & 0.3790 & 0.8923 & 0.4018 & \underline{0.9142} & \underline{0.1703} & \underline{0.8388} & \underline{0.1254} & \underline{0.9062} & 0.0676 & 0.9108 & 0.2469 \\ 
        PLE & 0.8466 & \underline{0.3769} & 0.8928 & 0.4007 & 0.9131 & 0.1713 & 0.8370 & 0.1265 & 0.9054 & 0.0678 & 0.9115 & \underline{0.2461} \\
        TAML & \underline{0.8467} & \textbf{0.3768} & \underline{0.8935} & \underline{0.3996} & 0.9132 & 0.1711 & 0.8297 & 0.1301 & \underline{0.9062} & \underline{0.0674} & \underline{0.9116} & \textbf{0.2459} \\
        STAR & 0.8431 & 0.3812 & 0.8900 & 0.4077 & 0.9121 & 0.1724 & 0.8325 & 0.1294 & 0.9061 & \textbf{0.0671} & 0.9095 & 0.2485 \\
        \midrule
        Ours & \textbf{0.8510} & 0.3875 & \textbf{0.8951} & \textbf{0.3983} & \textbf{0.9162} & \textbf{0.1688} & \textbf{0.8714} & \textbf{0.1164} & \textbf{0.9139} & 0.0677 & \textbf{0.9133} & 0.2500 \\
        RelaImpr. & 1.24\% & - & 0.41\% & - & 0.48\% & - & 9.62\% & - & 1.89\% & - & 0.41\% & - \\ 
    \bottomrule
    \end{tabular}
    \label{tab:main_res}
    }
\end{table*}

\subsubsection{Compared models} The following models are adopted to verify the effectiveness of our model.
\begin{itemize}
    \item \textbf{Single Task.} A simple MLP model trained only with the samples corresponding to the specified conversion action.
    \item \textbf{Shared Bottom.} The Shared Bottom model employs a shared embedding layers to capture the connections between different tasks. For each conversion target, an individual MLP is adopted to obtain the prediction.
    \item \textbf{STAR.} STAR\cite{star} utilize star topology networks, which consists of a shared centered MLP and independent MLPs per task. The final weights of the neural network layer are obtained by element wise multiplying the weights of shared MLP and scenario-specific MLP.
    \item \textbf{MMoE.} MMoE\cite{mmoe} models the connections between different tasks by sharing the embedding layer and a mixture of expert networks. Subsequently, an individual gating network gathers the output from the experts and feeds the representations into an MLP to obtain the final results for each task.
    \item \textbf{PLE.} PLE\cite{PLE} enhances MMoE by utilizing task-level experts and thus mitigating the negative transfer problem. Additionally, PLE stacks its CGC layers in order to capture deeper semantic representations.
    \item \textbf{TAML.} TAML\cite{TAML} further separates task-level experts into general task experts and learner task experts and utilizes self-distillation losses to make robust predictions.
\end{itemize}
\subsubsection{Implementation details}
For all the baselines and our models, we train the models in a multi-task style. A training batch includes data from multiple target conversion actions. All the baselines employ the base mask strategy to guarantee that sorely the gradient of the corresponding loss will be back-propagated to the weights while KAML utilize ADM to extend the masks. All models are trained with Adam\cite{Adam}, the learning rate is set to $0.001$ and the batch size is $5000$. Each experiment was conducted in five rounds, and the resulting data was averaged and reported in the paper.
\subsubsection{Metrics}
The widely used metrics Area Under the ROC Curve (AUC) and Logloss are deployed for evaluation. For AUC, a higher value indicates a superior performance, whereas for Logloss, a lower value is better. Additionally, one sample will be only evaluated once, which depends on its target conversion action. For example, for a sample which the advertiser pays for \textit{Registration}, only the label of \textit{Registration} will be used for evaluation, even though the advertiser reports a \textit{Activation} event. All the samples are evaluated separately according to the conversion actions. The overall performance are conducted on the entirety of the test set, with each sample being assigned its label by the target conversion action.

Besides, we provide relative improvements\cite{shen2021sar,yan2014coupled} for better comparison, which is based on AUC:
\begin{equation}
    \text{RelaImpr} = \frac{\text{AUC}(KAML)-0.5}{\text{AUC}(\text{BestBaseline})-0.5} -1 ~.
\end{equation}

\subsection{Offline Experiment}
\subsubsection{Performance on the industrial dataset}
We employ our model and all the baselines on the industrial dataset, whose results are depicted in Table \ref{tab:main_res}. From the results, we can conclude that:

\begin{itemize}
    \item The existing multi-task frameworks do not stably outperform SingleTask. Specifically, for Conversion Action A, SingleTask relatively improves about 0.89\% in AUC, compared with the best baseline TAML. This may be attributed to the negative transfer phenomenon, where the introduction of samples from other target conversion actions leads to additional noises.
    \item The results demonstrate that the performance of the various models differs significantly when evaluated across different conversion actions. STAR ranks the last over almost all tasks except the Action E. It implies that STAR is more suitable for the tasks involving a large number of scenarios or tasks, where the experiment of its original paper covers 19 scenarios. TAML and SharedBottom obtains good performances in some tasks respectively, which underscores the difficulty of the multi-task modelling. With regard to MMoE and PLE, their performance is relatively stable. In comparison to MMoE, PLE demonstrates superior performance in Conversion Action A, B, D and across the entire tasks, which validating the effectiveness of CGC layer in PLE.
    \item AUC and Logloss are not consistent. For example, in Action A, TAML and KAML obtain the best LogLoss and AUC, respectively. A higher AUC does not means a lower Logloss, which may due to the different architecture of models.
    \item KAML outperforms all the baselines in the view of individual tasks and the overall performance. Based on MMoE, KAML gains consistent improvements over MMoE, which substantiates the three modules we introduced in the paper. With regard to Logloss, we find that KAML exhibits a certain degree of instability. Specifically, KAML does not achieve the optimal  Logloss in Conversion Action A and Action E. This phenomenon may be attributed to the RLU module, as reported in other work\cite{sheng2023joint,lin2024understanding}. It has been demonstrated that ranking loss can result in a decline in Logloss while concurrently enhancing AUC. The detailed analysis of the RLU module will be presented in the ablation study.
\end{itemize}

\subsubsection{Performance on the simulated public dataset}

In order to emulate the phenomenon that advertisers only submit a portion of user conversion actions, on the public dataset KuaiRank-Pure, three groups of experiments are conducted:
\begin{itemize}
    \item \textbf{Oracle}. All labels for all the feedbacks are available and used for training, which corresponds to the advertisers submitting all the conversion actions.
    \item \textbf{Vanilla}. For all samples, as mentioned above, only one type of feedback is randomly selected as the target conversion action. Only the feedbacks corresponding to the simulated target conversion actions are used for training.
    \item \textbf{KAML}. In addition to the correlation feedbacks, two other feedbacks are randomly adopted for training, which approximates the mask strategy in ADM of KAML. Furthermore, rank loss is used to enhance the ability of the model. Given that the conversion actions in the dataset are randomly selected, the issue of distribution discrepancy is rendered moot. Consequently, HKE is not introduced in this context here.
\end{itemize}

As demonstrated in Table \ref{tab:public_res}, the results of the simulated public dataset reveal that, in three out of four conversion actions, KAML outperforms Vanilla. This outcome verifies the effectiveness of RLU and ADM modules in KAML. Surprisingly, KAML is even better than Oracle in Action C and Action D. This phenomenon may be attributed to the seesaw effect, wherein the introduction of additional labels of Action A and Action B can compromise the performance of Action C and Action D. 

\begin{table}[t]
    \centering
    \caption{Results on the simulated public dataset}
    \resizebox{\linewidth}{!}{
    \begin{tabular}{ccccccccc}
    \toprule
    Actions & \multicolumn{2}{c}{Action A} & \multicolumn{2}{c}{Action B} &
    \multicolumn{2}{c}{Action C} &
    \multicolumn{2}{c}{Action D} \\
        \cmidrule{1-9} Metric & AUC & LogLoss & AUC & LogLoss & AUC & LogLoss& AUC & LogLoss\\
        \midrule
        Vanilla & 0.7921 & 0.2724 & 0.6651 & 0.0299 & 0.7251 & 0.0401 & 0.6585 & 0.0168 \\
        KAML & 0.8523 & 0.1392 & 0.6414 & 0.0258 & 0.7711 & 0.0371 & 0.7562 & 0.0257 \\
        \midrule
        Oracle & 0.9020 & 0.0959 & 0.6693 & 0.0367 & 0.7696 & 0.0407 & 0.7208 & 0.0296 \\ 
    \bottomrule
    \end{tabular}
    \label{tab:public_res}
    }
\vspace{-0.15in}
\end{table}

\subsubsection{Ablation Study}
\begin{table*}[t]
    \centering
    \caption{Results of Ablation Study}
    \vspace{-1pt}
    \resizebox{0.9\linewidth}{!}{
    \begin{tabular}{ccccccccccccc}
    \toprule
    Conversion Action & \multicolumn{2}{c}{Action A} & \multicolumn{2}{c}{Action B} &
    \multicolumn{2}{c}{Action C} &
    \multicolumn{2}{c}{Action D} &
    \multicolumn{2}{c}{Action E} &
    \multicolumn{2}{c}{All} \\
        \cmidrule{1-13} Metric & AUC & LogLoss & AUC & LogLoss & AUC & LogLoss& AUC & LogLoss& AUC & LogLoss& AUC & LogLoss\\
        \midrule
        MMoE & 0.8448 & 0.3787 & 0.8921 & 0.4024 & 0.9139 & 0.1706 & 0.8300 & 0.1275 & 0.9061 & 0.0675 & 0.9108 & 0.2469 \\ 
        MMoE+ADM & 0.8473 & 0.3763 & 0.8930 & 0.4002 & 0.9138 & 0.1710 & 0.8599 & 0.1218 & 0.9118 & 0.0658 & 0.9122 & 0.2450 \\
        MMoE+ADM+HKE & 0.8490 & 0.3746 & 0.8946 & 0.3972 & 0.9165 & 0.1692 & 0.8677 & 0.1177 & 0.9121 & 0.0657 & 0.9133 & 0.2437 \\
        MMoE+RLU & 0.8453 & 0.3794 & 0.8928 & 0.4019 & 0.9139 & 0.1723 & 0.8486 & 0.1227 & 0.9078 & 0.0723 & 0.9102 & 0.2489 \\
        KAML & 0.8510 & 0.3875 & 0.8951 & 0.3983 & 0.9162 & 0.1688 & 0.8714 & 0.1164 & 0.9139 & 0.0677 & 0.9133 & 0.2500 \\
    \bottomrule
    \end{tabular}
    \label{tab:abalation_res}
    }
    \vspace{-0.1in}
\end{table*}
To validate the three modules proposed in the paper, a series of ablation studies are conducted. Based on MMoE, we compare KAML with the following alternatives:
\begin{itemize}
    \item \textbf{MMoE.} We adopt MMoE as the base model of KAML.
    \item \textbf{MMoE+ADM.} MMoE with Attribution Driven Masking.
    \item \textbf{MMoE+ADM+HKE.} MMoE with Attribution Driven Masking and Hierarchical Knowledge Extraction. It is important to note that HKE cannot be utilized without ADM, given that it employs extended samples to train the sub-towers.
    \item \textbf{MMoE+RLU.} MMoE with Ranking-based Label Utilization.
\end{itemize}
The overall result is shown in Table \ref{tab:abalation_res}. For the comparision of MMoE+ADM with MMoE, as illustrated in Table \ref{tab:dataset_count}, a markedly greater number of samples are employed for training of each individual task. The comparison of these two methods on AUC and LogLoss also reveals that the mask strategy of the ADM exactly filters out noisy labels and provides the model with more appropriate samples, thereby enhancing its performance. Furthermore, the incorporation of HKE serves to further improve the performance, which proves that the design of hierarchical towers effectively addresses the distribution discrepancy issue previously highlighted. 

When turns to RLU, things seems to be a little different. Compared with MMoE, MMoE+RLU outperforms in all the scenarios except the Scenario 3 in AUC. Additionally, Logloss of MMoE+RLU is worse in 3 of all the 5 scenarios, though AUC are higher. It may due to that the optimization objective of ranking losses does not ensure a lower Logloss, which is also reported in other work\cite{sheng2023joint,lin2024understanding}.


\subsection{Online Experiment}
\subsubsection{Online Deployment}
To verify the effectiveness of KAML, we deploy KAML on Huawei's online advertising platform for CVR prediction. More than 100 features, including sparse features and dense features are used for the model. We conduct a 8-day online A/B test on the online advertising platform, where a well-crafted MTL model is selected as the baseline. For online serving, both the baseline and our model are trained on the same latest exposure logs to ensure comparability. As for the traffic proportion, both baseline and KAML serve approximately 10\% of the total traffic.

\subsubsection{Performance}
\begin{figure}[t]
	\centering
        \subfigure[RPM]{
		\label{fig:kctl_online_rpm}
		\includegraphics[width=0.48\linewidth]{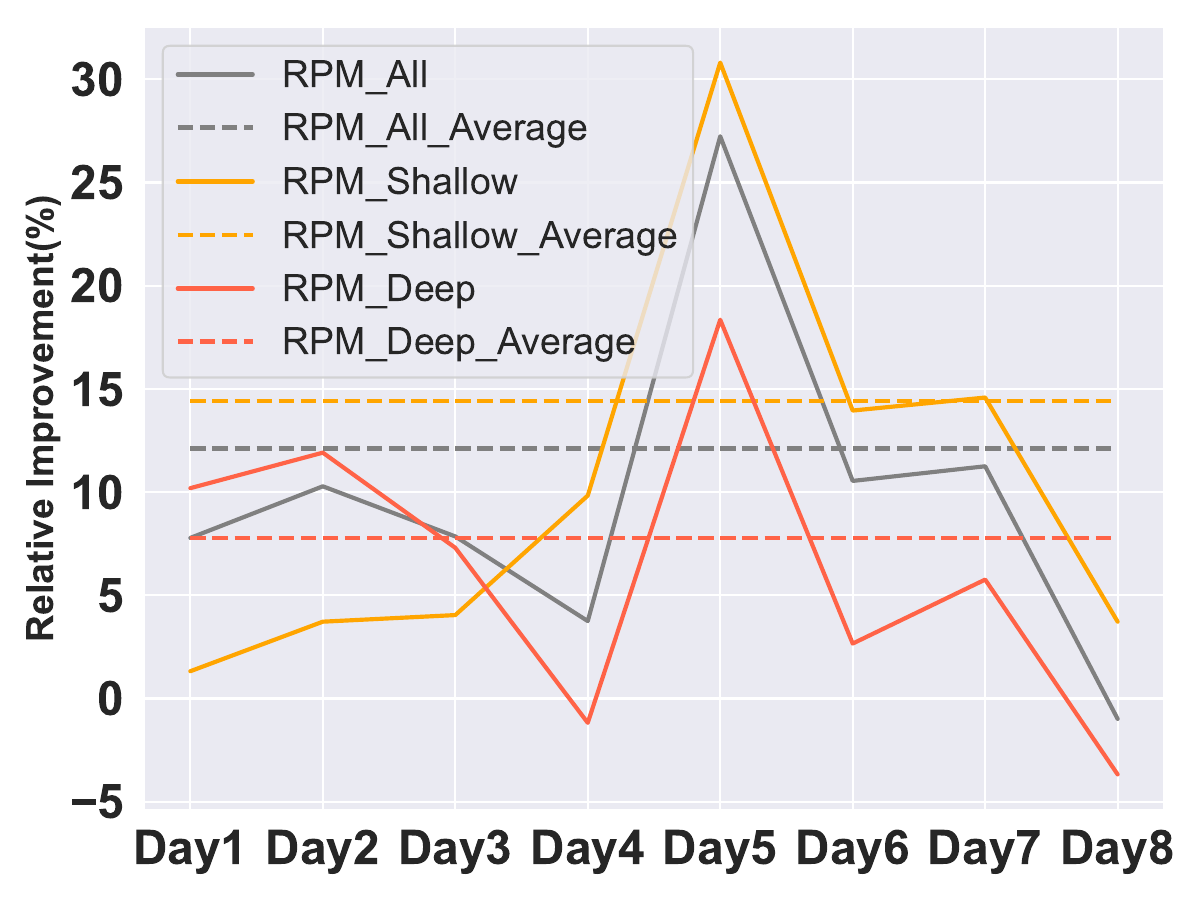}}
        \subfigure[CVR]{
		\label{fig:kctl_online_cvr}
		\includegraphics[width=0.48\linewidth]{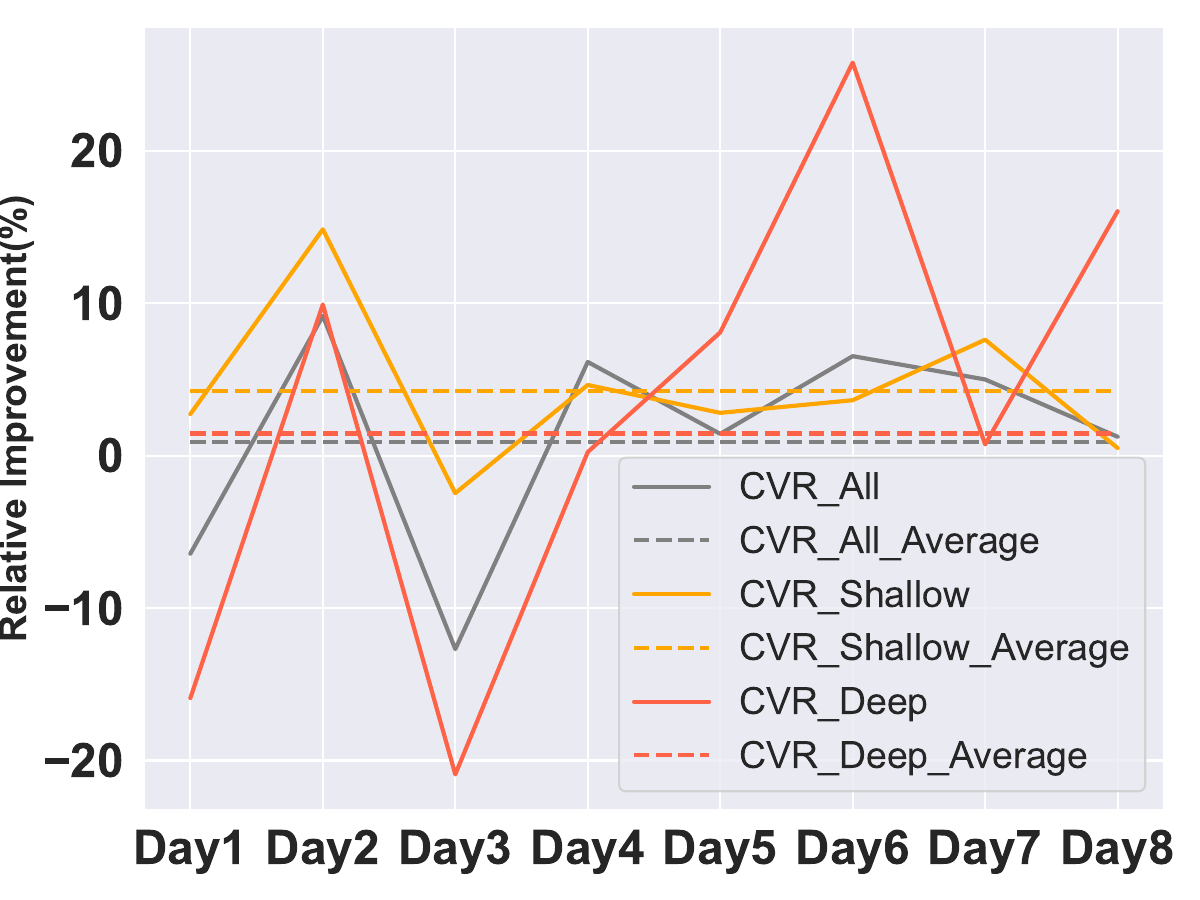}}
\vspace{-10pt}
\caption{Result of Online A/B test. The conversion actions are divide into two groups: Shallow and Deep by their specific conversion actions.}
\label{fig:kctl_online}
\vspace{-8mm}
\end{figure}

The online result is shown in Figure \ref{fig:kctl_online}. The online performances are presented from two distinct perspectives. The key performance indicators are RPM (Revenue Per Mille) and CVR. According to the conversion action, we divide the curve of improvements into three groups: Shallow (Re-engagement and Activation), Deep(Paymeny, Registration and Retention) and Overall. The figure demonstrate that our model obtains stable improvements in both RPM and CVR. On average, KAML gains 12.11\% and 0.92\% improvements on RPM and CVR, respectively.
\section{RELATED WORK}

Conventional CVR modeling methods typically build upon techniques developed for CTR prediction tasks, evolving from traditional shallow models\cite{shallow1, shallow2} to more complex deep learning approaches as deep learning has proven highly effective\cite{deepfm,autoint,dfm_v0,dfm, Apppromotion}.
However, issues such as data sparsity\cite{dqy, ranking}  and the complex relationships\cite{esmmv2, dbmtl} between conversion tasks make CVR modeling particularly challenging. 
In real-world applications, multi-task learning (MTL) is widely adopted to jointly train a unified model\cite{likemtl, surveyMTL, chen2020just, yang2023adatask} that can handle related tasks simultaneously. These multi-task recommendation models typically adopt an architecture composed of four main modules: Expert, Gate, Tower, and tailored loss functions. The Expert module extracts knowledge from the input features embedding in multi-view manner. The Gate module dynamically controls the weights of each expert and aggregates their outputs. The Tower module then uses the combined expert outputs to predict task-specific probabilities. Finally, tailored loss functions guide the learning process to optimize performance across tasks.

This architecture enables effective knowledge sharing and transfer across tasks. To enhance knowledge sharing and alleviate conflicts among tasks, various models have been proposed. Since complex problems may contain many sub-problems each requiring different experts, several MoE\cite{moe} models have been developed. 
MMoE\cite{mmoe} extends MoE  by incorporating multiple gates for each task, enabling it to learn distinct fusion weights that combine shared experts, thereby modeling task relationships more effectively. Zhao et al.\cite{mmoeplus} further explored soft-parameter sharing techniques, including MMoE, to optimize multiple ranking objectives in video recommendation. PLE\cite{PLE} goes a step further by explicitly separating task-shared and task-specific experts to better address the seesaw phenomenon. 
Building on this framework, Liu et al.\cite{TAML} propose a multi-learner network to replace the single tower, generating scores by fusing outputs from multiple learners.
Xi et al.\cite{AITM} consider the sequential dependencies among multi-step conversions and establish relationships between tower modules, thereby capturing dependencies across conversion steps.

Another approach to improving MTL performance is to design specialized loss functions that guide the learning process to optimize performance across tasks. ESMM\cite{esmm} leverages the relationship between CTR and CVR by modeling their product to estimate the post-impression click-through conversion rate (CTCVR).
CTnoCVR\cite{ctnocvr} introduces a loss function that penalizes instances where clicks do not lead to conversions.
Additionally, Multi-IPW\cite{cmtl} and ESCM2-IPW\cite{esmm2} employ causal inference methods to design customized loss functions that mitigate the effects of click propensity, thus improving task accuracy and robustness.

In addition to these approaches, various other efforts have been made to enhance CVR prediction, such as designing multi-embedding mechanism\cite{STEM}, modeling  CVR across multiple domains \cite{zhao2024retrievable, star, su2022cross, huan2023samd}, and integrating multi-domain recommendation with multi-task recommendation\cite{mask, zhang2024m3oe, zhang2022leaving}. 
However, these methods are not specifically designed to address asymmetric multi-label challenges and fall short in effectively modeling all conversion tasks.

\section{CONCLUSION}
In this paper, we introduced a novel framework, KAML, to relieve the issue of incomplete labels and train a unified model with the incomplete and skewed multi-label data for conversion rate prediction.
Specifically, we proposed an attribution-driven masking (ADM) strategy to efficiently utilize the asymmetric multi-label data. 
To address the distribution discrepancy problem, hierarchical knowledge extraction mechanism (HKE) are proposed.
Besides, the incorporation of ranking loss further enhances the model's ability to learn from sparse and incomplete data.
Comprehensive offline and online experiments validate the effectiveness of our framework and KAML is currently serving as the main framework in the CVR prediction stages of Huawei's online advertising platform.




\bibliographystyle{ACM-Reference-Format}
\bibliography{my_ref}










\end{document}